\documentclass[a4paper]{article}

\usepackage[english]{babel}
\usepackage[utf8]{inputenc}
\usepackage{amsmath}
\usepackage{graphicx}
\usepackage[colorinlistoftodos]{todonotes}
\usepackage{amssymb,amsthm}

\title{Interactive Image Manipulation \\ with Natural Language Instruction Commands}

\author{
  Seitaro Shinagawa*1*2\\
  \texttt{shinagawa.seitaro.si8@is.naist.jp} \\
  \and
  Koichiro Yoshino*1*3 \\
  \texttt{koichiro@is.naist.jp} \\
  \and
  Sakriani Sakti*1 \\
  \texttt{ssakti@is.naist.jp} \\
  \and
  Yu Suzuki*1 \\
  \texttt{ysuzuki@is.naist.jp} \\
  \and
  Satoshi Nakamura*1*2 \\
  \texttt{s-nakamura@is.naist.jp} \\
  \and
  *1 Nara Institute of Science and Technology \\
  *2 RIKEN, Center for Advanced Intelligence Project AIP \\
  *3 PRESTO, Japan Science and Technology Agency \\
}

\date{}

\begin{document}
\maketitle


\begin{abstract}
We propose an interactive image-manipulation system with natural language instruction, which can generate a target image from a source image and an instruction that describes the difference between the source and the target image. The system makes it possible to modify a generated image interactively and make natural language conditioned image generation more controllable. We construct a neural network that handles image vectors in latent space to transform the source vector to the target vector by using the vector of instruction. The experimental results indicate that the proposed framework successfully generates the target image by using a source image and an instruction on manipulation in our dataset.
\end{abstract}

\section{Introduction}
\label{sec:intro}
\vspace{-1mm}
Specialized skills are required to create a commercially available image. One way to obtain a required image at low cost is by finding existing images through an image search. However, it is difficult to obtain what was exactly imagined since the desired image may not exist on the Web. An automatic image-generation system using natural language has the potential to generate what was actually imagined without requiring any special skill or cost. 
The solution to this challenging task should address not only practical benefits but also contributions of bridging natural language understanding with image processing.
Image generation task from natural language have been investigated as ``cap2image'' and several deep neural network (DNN)-based generative models are successful \cite{mansimov2016,reed2016b}. 
Although it is difficult to define the relationships between languages and images clearly, DNN-based models make it possible to align these relationships in the latent space. The network is composed of {\em language-encoder} and {\em image-decoder}. Long short-term memory (LSTM) \cite{hochreiter1997long} is generally used as a {\em language-encoder}, and several network structures; Variational auto-encoder \cite{kingma2014auto}, generative adversarial network (GAN) \cite{goodfellow2014generative}, and pixelCNN \cite{van2016conditional} are used as an {\em image-decoder}. 

As far as our knowledge, Reed et al.~\cite{reed2016b} firstly succeeded to construct a discriminable image generator conditioned by a caption based on a deep convolutional generative adversarial network (DCGAN) \cite{radford2015unsupervised}. We start at this work from a different viewpoint; we focus on a practical problem in this task. 
It is possible that they generate a slightly different image from what the user actually wanted. Our motivation is to tackle this point by introducing an interactive manipulation framework and make a generated image modifiable with natural language. 



\begin{figure}[t]
\vspace{-5mm}
\centering
\includegraphics[width=1.0\textwidth]{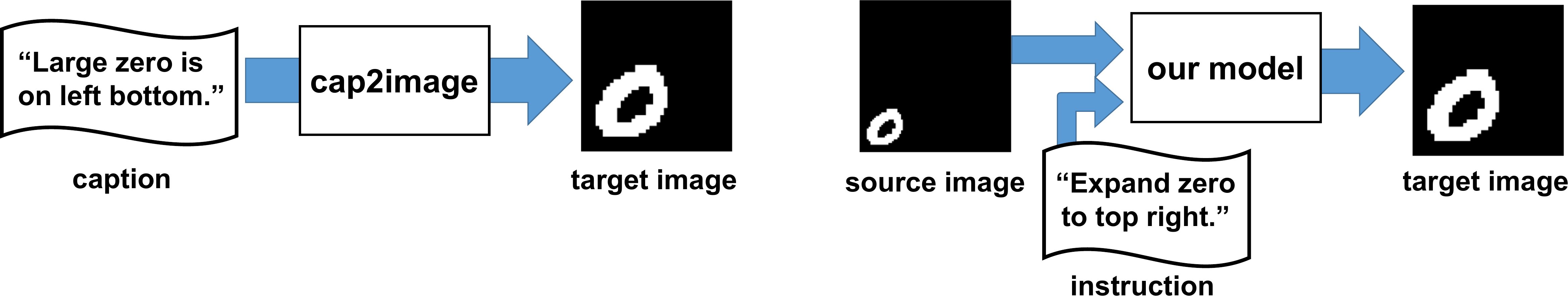}
\vspace{-5mm}
\caption{\label{fig:frameworks}Comparison of natural language conditioned image generation framework between cap2image (left) and the proposed framework (right).}
\vspace{-4mm}
\end{figure}

Figure~\ref{fig:frameworks} shows the difference of the cap2image framework and the proposed framework. 
Compared with the cap2image framework models, the proposed framework model generates a new image from the source image and the instruction that represents the difference between the source and the target image. 
In our insight, the advantage of the proposed framework is to allow users to modify the source image that has been generated. Furthermore, users only have to focus on the difference and represent it as natural language. It is not only easier for user to use but also easier for {\em language-encoder} to learn because the instruction with a few  difference information will be much shorter than caption with all information of the desired image.
We define a latent space composed of image feature vectors and set a problem of manipulation as a transformation of a vector in latent space. The manipulated image is generated from the latent vector that is transformed from the latent vector of the original image by the embedded natural language instruction. 

Kiros et al.~\cite{kiros2014unifying} reported that it is possible to learn a model whose shared latent space between languages (captions) and images in a DNN has the characteristic of additivity. Reed et al.~\cite{reed2015} reported that it is possible to generate the target image using image analogy. 
According to these property, we realize the image manipulation system by bridging the analogy in the latent space of image and natural language instruction, as $\{source\:image\}+``instruction"=\{target\:image\}$.
We confirm that there are many related works to edit image flexibly using user hand-drawing \cite{brock2016neural,zhu2016generative}. However, manipulating images with natural language will be useful to get a moderate image easily if we can bridge the natural language and modification which contains many drawing operations.



\section{Network architecture of proposed framework}
\vspace{-1mm}
The network architecture of the proposed framework to generate an image from a source image and a language instruction is shown in Figure~\ref{fig:fine-tune}. The framework is composed of an encoder and a decoder as existing image generators.
Details of the encoder and decoder models are described in this section. 

\vspace{-1mm}
\subsection{Encoder model}
\vspace{-1mm}
The encoder model consists of two parts, an image encoder $ImEnc$ and instruction encoder $IEnc$. In the image encoder, we use the same architecture as the discriminator of a DCGAN; In the instruction encoder, we use a plain LSTM \cite{hochreiter1997long} without peephole. 
We assume that a source image is $x_{im}$ and instruction text sequence is $S=[s_1,s_2,\cdots,s_T]$. Then each encoder transformation is defined as,

\vspace{-5mm}
\begin{eqnarray}
  \phi_{im} &=& CNN_{ImEnc}(x_{im}) \\
  \phi_{i}^{t} &=& LSTM_{IEnc}(s_{t},\phi_{i}^{t-1}) \quad (where,  \phi_{i}^{0}=\bf{0})\\
  \phi_{fc} &=& FC(\phi_{im},\phi_{i}^T)
\end{eqnarray}
$\phi_{im}$ represents the source image vector. $\phi_{i}^{t}$ is the hidden vector of $ImEnc$ in the time step $t$, then $\phi_{i}^T$ represents the instruction vector. $\phi_{im}$ and $\phi_{i}^T$ are both fed into one free connected ({\sf FC}) layer, 
and the output $\phi_{fc}$ is trained to be the latent variable of the target vector. If $\phi_{fc}$ can be the latent variable of the target image through learning, the learned model can generate target images without modifying the DCGAN proposed by Reed et al.~\cite{reed2016b}. We used a single layer for the {\sf FC} layer because we assumed that images in latent space can be transformed linearly, as reported in Kiros et al.~\cite{kiros2014unifying}, and we would like to align language instruction to the linear transformation of images with single non-linear transformation.

\vspace{-1mm}
\subsection{Decoder model} 
\begin{figure}[t]
\vspace{-5mm}
\begin{minipage}{0.5\hsize}
\centering
\includegraphics[width=1.0\textwidth]{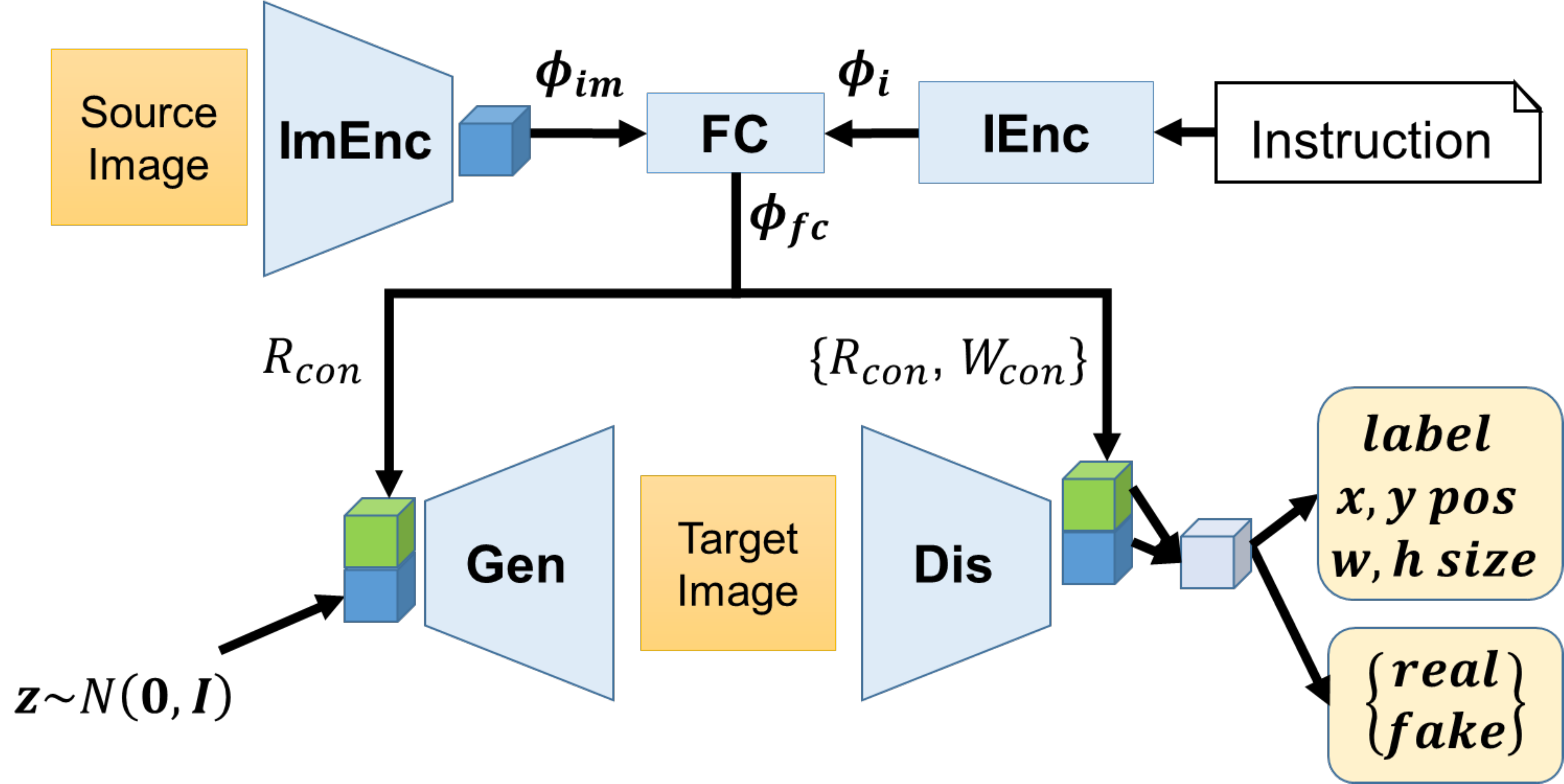}
\vspace{-5mm}
\caption{\label{fig:fine-tune}Architecture of proposed framework}
\vspace{-5mm}
\label{fig:one}
\end{minipage}
\begin{minipage}{0.5\hsize}
\centering
\vspace{3mm}
\includegraphics[width=0.9\textwidth]{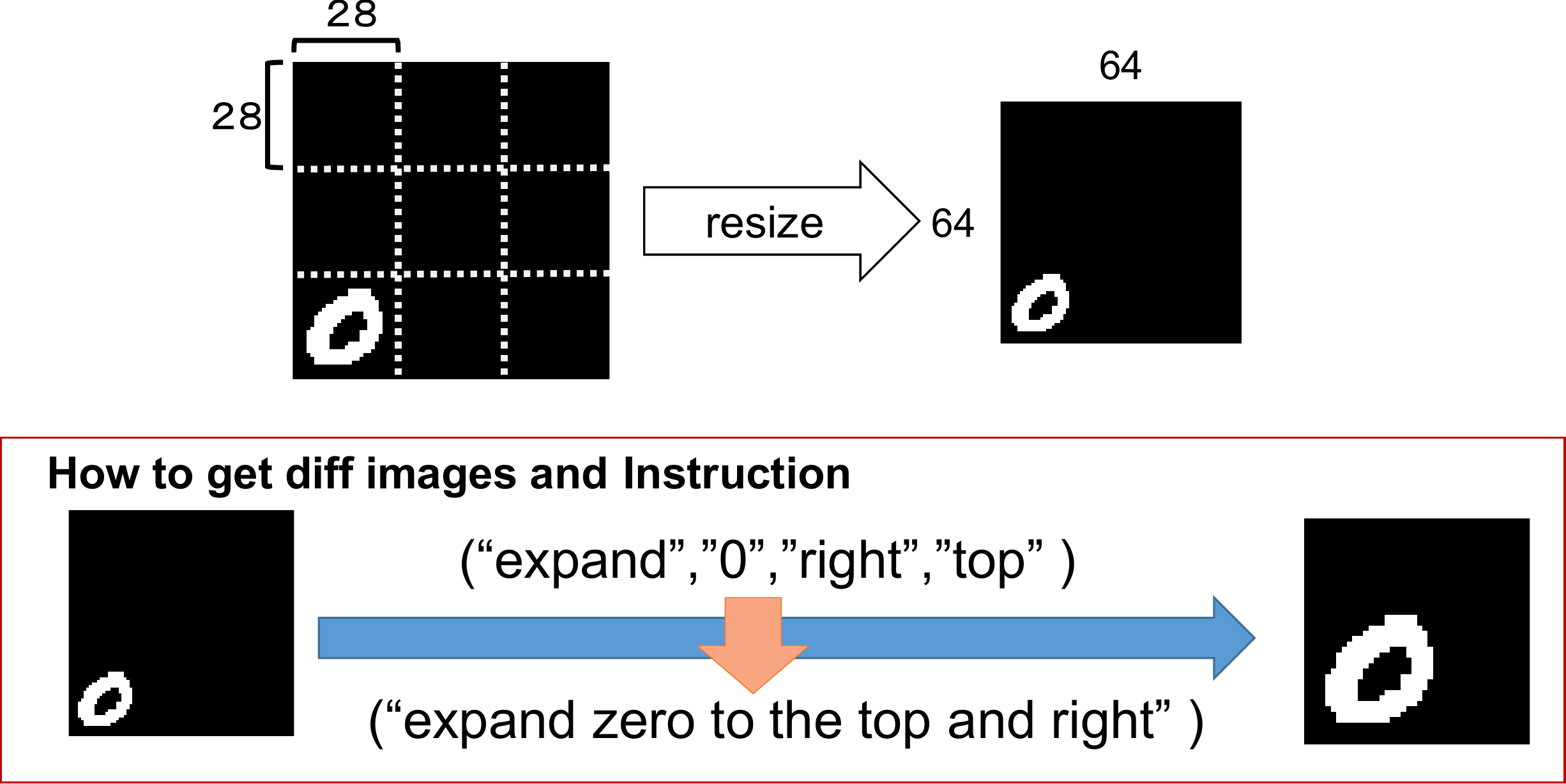}
\vspace{-1mm}
\caption{\label{fig:amnist}Artificial MNIST}
\label{fig:two}
\vspace{-5mm}
\end{minipage}
\end{figure}

\vspace{-1mm}
In the decoder model, we basically use the same DCGAN as Reed et al.~\cite{reed2016b} (Figure~\ref{fig:fine-tune}); however, the final layer-activation function of our generator is linear because it was necessary to succeed model training in our trials. In this setting, the range of generated image-pixel values is unlimited. We clipped the values in the range $[0,1]$ because some pixel values are out of the range of the training data, which have the range $[0,1]$. We used class labels, object positions, and size labels by following Odena et al.~\cite{odena2016conditional} to stabilize the training instead of using latent-space interpolation that Reed et al.~\cite{reed2016b} proposed. This is because the label information was essential for training in our experience. While training our model, we used feature matching \cite{salimans2016improved} to stabilize GAN training. 


\section{Experiments}
\subsection{Dataset}
\label{Sec:Dataset}
For the experiment, we constructed a dataset of images controlled using natural language instruction by using MNIST \cite{lecun1998gradient} dataset and manually created modifications. 
The main reason to use artificial data is that we want to analyze the learned model.
This setting also makes it easy to collect a lot of examples. 
Figure~\ref{fig:amnist} shows an example of data in the corpus. To construct the data, we prepared a canvas that was three times larger than that of the original MNIST data. We also prepared an instruction verb set, \{``put",``remove",``expand",``compress",\\``move"\}, position set, \{``top",``left",``right",``bottom",``middle"\}, and direction set \{``top",``left",``right",``bottom",``top left",``top right",``bottom left",``bottom right"\}  to create instructions. The simulator determined a triplet of instructions, ``verb'', ``digit class'', and ``position'', as shown in Figure~\ref{fig:amnist}, and created a transformed image according to this triplet. Instructions were also automatically generated using the triplet. Each canvas had image, digit-class (11-class), position $(x,y)$ (\{3,3\}-class), and size $(width,height)$ (\{4,4\}-class) information. None of the digit objects had another class, $(x,y)=(0,0)$,$(width,height)$=(0,0). Thirty-one unique images for one-digit data in the MNIST were generated. In total, there were 369 triplets of source image, target image, and instruction for each train/test sample.

\subsection{Experimental setting}
We used 1000 samples on each $0 \sim 9$ class from the original MNIST training set with 60,000 samples, then obtained 10,000 samples in the dataset. We prepared 3,690,000 triplets in accordance with the data preparation described in Section~\ref{Sec:Dataset}. We divided them into {training: 90\% and validation: 10\%}. For the test, we used another 100 samples on each $0 \sim 9$ class from the original MNIST, we obtained 1,000 samples for testing. We used Chainer\footnote{http://chainer.org/}\cite{chainer2015} for the implementation. We used the following conditions: images are resize to 64x64, latent-space dimension $=128$, optimization $=Adam$ \cite{kingma2014adam} (initialized by $\alpha = 2.0 \times 10^{-4},\beta = 0.5$), and training-time epochs = 20.

We evaluated the generated image by comparing it to the target image. We used structural similarity (SSIM) (higher is better) \cite{wang2004image}, as in Mansimov et al.~\cite{mansimov2016}, to measure the similarities between the target and the generated images. 


\section{Results}
\begin{figure}[t]
\vspace{-5mm}
\centering
\includegraphics[width=1.0\textwidth]{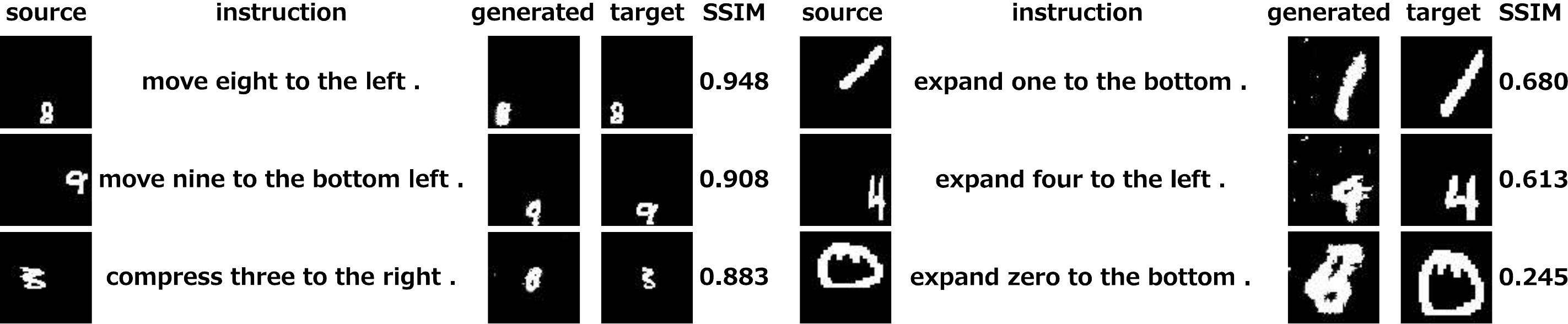}
\vspace{-4mm}
\caption{\label{fig:result_example} Examples generated with our framework. Examples are randomly sampled from top 10\%, 10\%-20\% and 20\%-30\% (left) and bottom 30\%-20\%, 20\%-10\% and 10\% (right) groups in SSIM.}
\end{figure}

Figure~\ref{fig:result_example} shows examples generated with the proposed framework. Source images and instructions (first and second columns) were given to the model. The generated images, the target (gold) images, and the SSIMs are shown in the third to fifth columns. From these examples, we could confirm that our framework can generate similar images to the target images, especially regarding positions and sizes. 
We conducted a subjective evaluation by three subjects. Each subject evaluated similarities of generated images and target (gold) images with 5 degrees (5=very similar, 1=very different).
The subjective evaluation is composed of 100 example following Figure~\ref{fig:result_example} format without SSIM. The  rate of the score was \{1:~9.00\%, 2:~10.7\%, 3:~18.3\%, 4:~16.7\%, 5:~45.3\%\}. 

\begin{figure}[t]
\centering
\vspace{-4mm}
\includegraphics[width=1.0\textwidth]{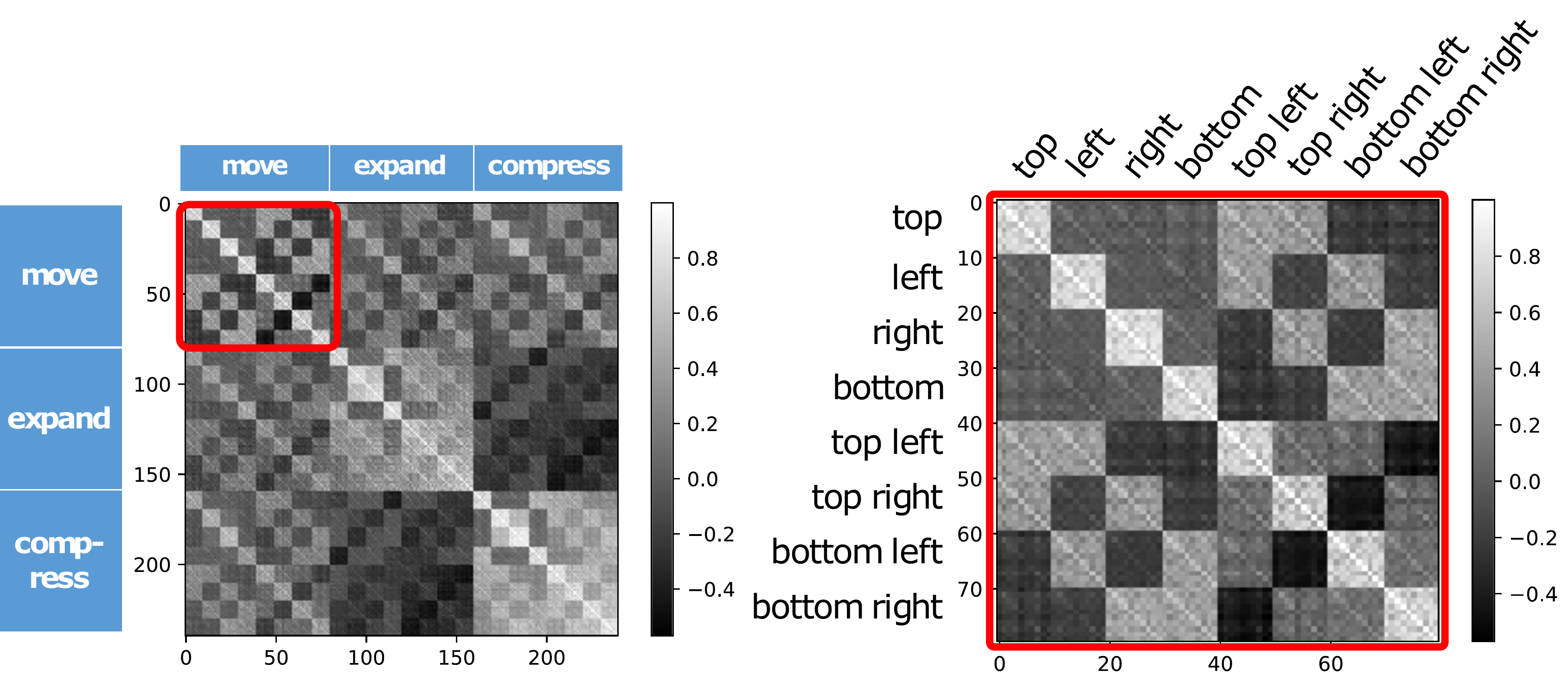}
\vspace{-6mm}
\caption{\label{fig:insvector} A visualization of instruction vector for ``move,'' ``expand'' and ``compress''. The left image shows a cosine similarity map. Each element shows the cosine similarity between two of instructions vector. The order is sorted by verb-direction-number. The right image shows the enlarged part (red squared) of ``move" instructions of the left image.}
\vspace{-6mm}
\end{figure}

Figure~\ref{fig:insvector} shows a visualization of the cosine similarities of instruction vectors of ``move,'' ``expand'' and ``compress''. They are sorted by verb-direction-number. The map is clearly separated by a certain size of blocks. 
The large black of ``expand"-``compress" in the left figure indicates that ``expand" and ``compress" are learned as the inverse. Furthermore, in the block of ``move"-``move" (the right figure), the enlarged part of the red square of the left figure, is also clearly separated by small blocks and the cosine similarities follow the direction similarity as well. These results indicate that instruction vectors learned the concept of verb and direction. However, the concept of number is not significant in the instruction vectors. We guess that it is because we used just one digit operation in the experiment. 
We also tried the visualization of ``put" and ``remove," but the clear blocks did not appear. 
We guess that this is because the concept of position or number is learned independently. 

We also tried inputting unseen operation, e.g.``move zero to the right" to the source image that has a digit zero in the right position, to investigate the limitation of our model. 
If our model learned the concept of instruction ideally, the zero should go away from the canvas, however, the digit did not go away from the canvas.
This is probably caused by that there are no instructions to take a digit away from the canvas except ``remove" in our dataset. 


\section{Discussion}
We proposed an image-manipulation framework using natural language instruction to make image generation systems more controllable. 
The experimental results indicate that the embedded instructions capture the concepts of operation apparently except for the digit information.
Our framework worked well for limited types of instructions. 
The results also indicate the potential to bridge the embedded natural language instructions and analogies of two images using our framework. 
Future work includes applying this framework to data that have a variety of images and manipulations.

\bibliographystyle{plain}

\clearpage
\appendix

\section{Avatar image manipulation}
In this section, we show an additional experiment that uses avatar images and natural language instructions from human collected by crowdsourcing as a more natural task setting.

\begin{figure}[h]
\begin{minipage}{0.5\hsize}
\centering
\vspace{0mm}
\includegraphics[width=1.0\textwidth]{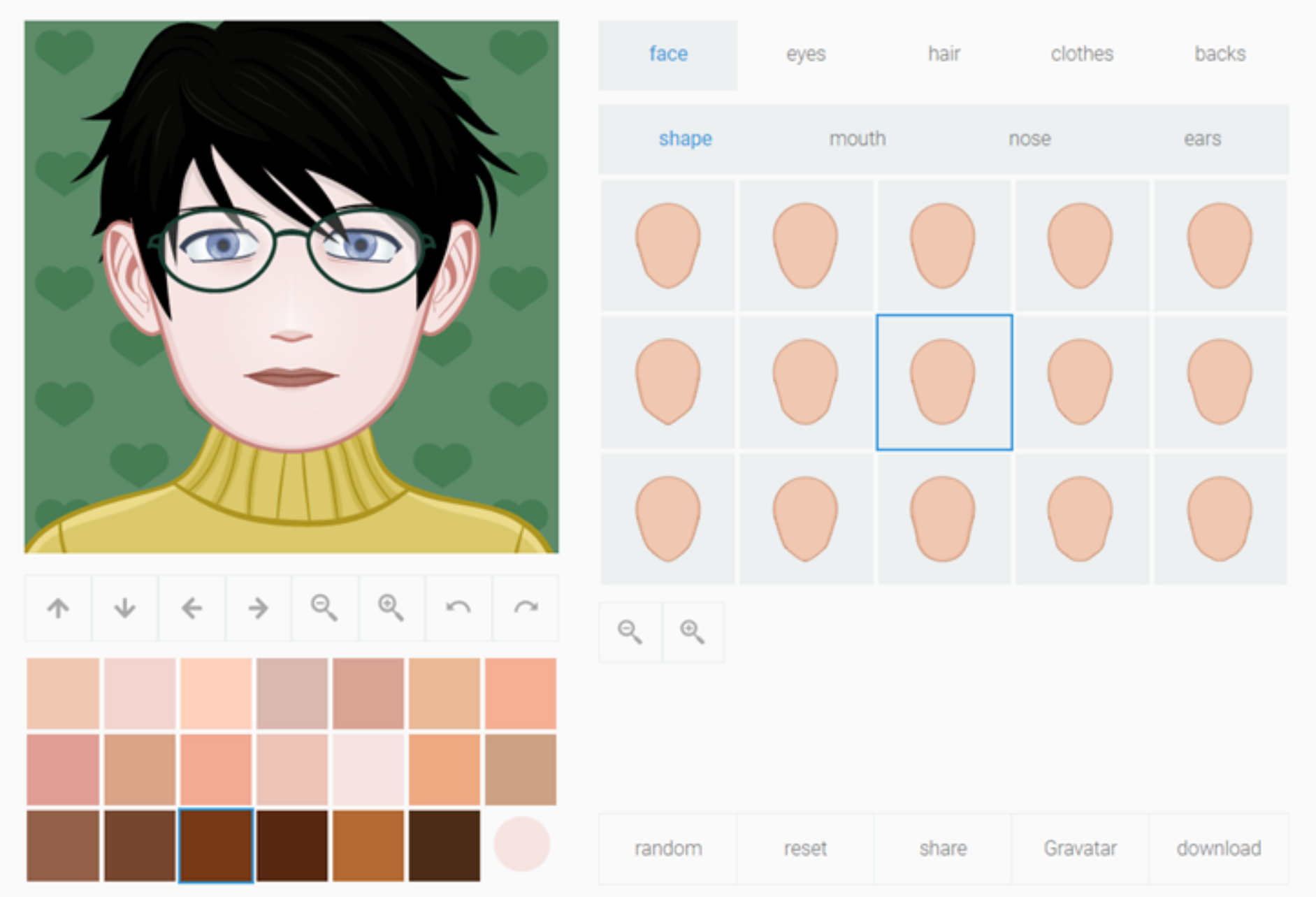}
\vspace{-2mm}
\caption{\label{fig:-1}System on AvatarMaker.com.}
\vspace{0mm}
\end{minipage}
\begin{minipage}{0.5\hsize}
\centering
\vspace{0mm}
\includegraphics[width=0.9\textwidth]{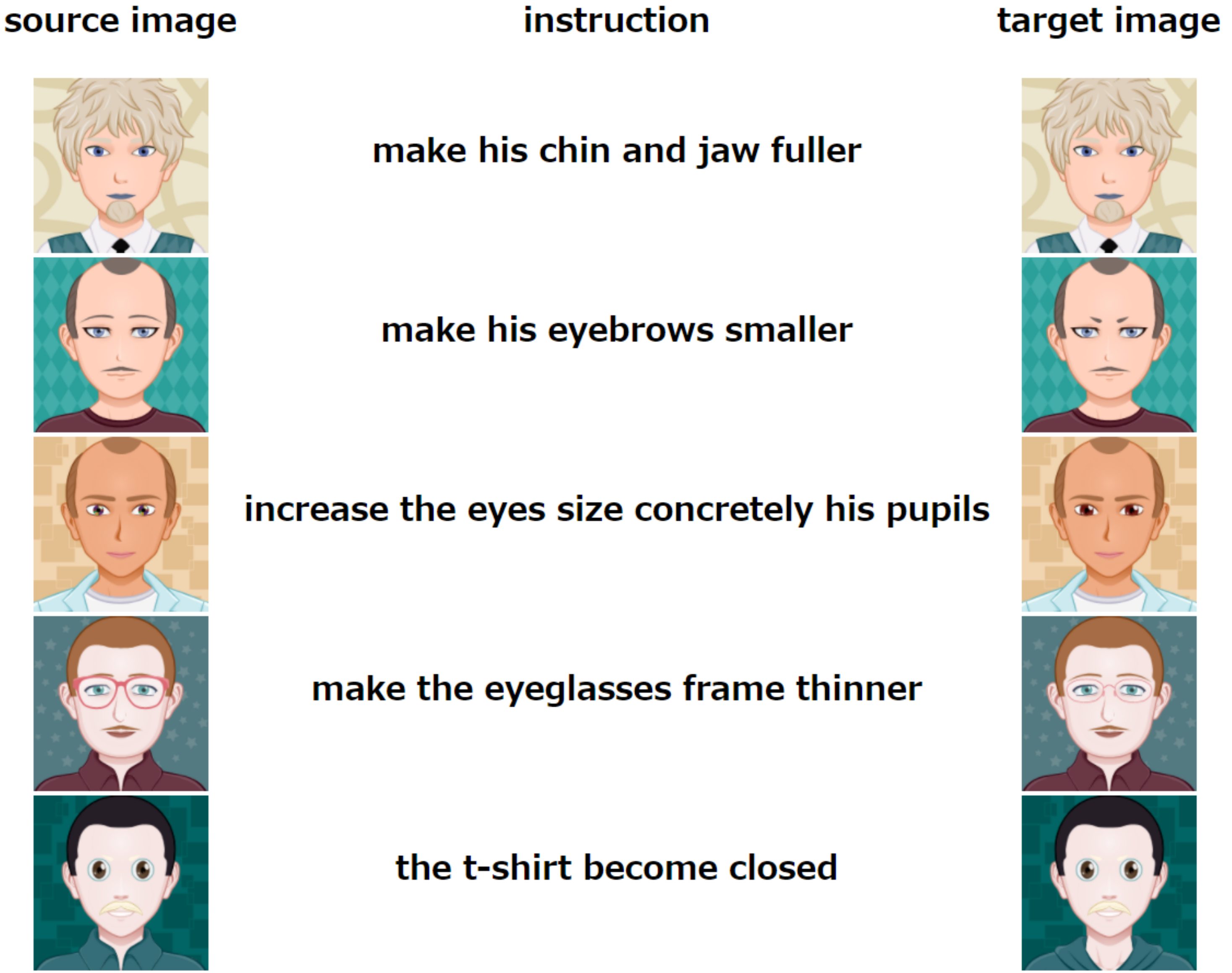}
\vspace{0mm}
\caption{\label{fig:0}Collected triplet of the avatar data.}
\vspace{0mm}
\end{minipage}
\end{figure}

\subsection{Data Collection}
We collected free avatar images on AvatarMaker.com \footnote{http://avatarmaker.com/}. Fig~\ref{fig:-1} shows the creating page of an avatar image. There are 14 kinds of attributes as follows:
\begin{itemize}
\item Male, Female
\item Face (shape:15, mouse:15, nose:15 ears:7)
\item Eyes (eye shape:15, iris:10, eyebrows:15, glasses:18)
\item Hair (on head:18, mustache:13, beard:13)
\item Clothes (13)
\item Backs (15)
\end{itemize}
 We randomly generated 8,000 images pair as source images.
 We randomly changed one attribute of them to generate target images and added one instruction that describes the differnce of them by using crowdsourcing (Fig~\ref{fig:0}).

\subsection{Experimental settings}
We tried a simple semi-supervised setting, because our data-set is still small. We divided the annotated \{source image, target image, instruction\} triplet into 7,000 for training, 500 for validation and test, respectively. In the training, We used supervised learning and unsupervised learning alternatively. The unsupervised learning is realized by the instruction vector to be a zero vector. 

\subsection{Results}
\begin{figure}[t]
\centering
\vspace{-10mm}
\includegraphics[width=1.0\textwidth]{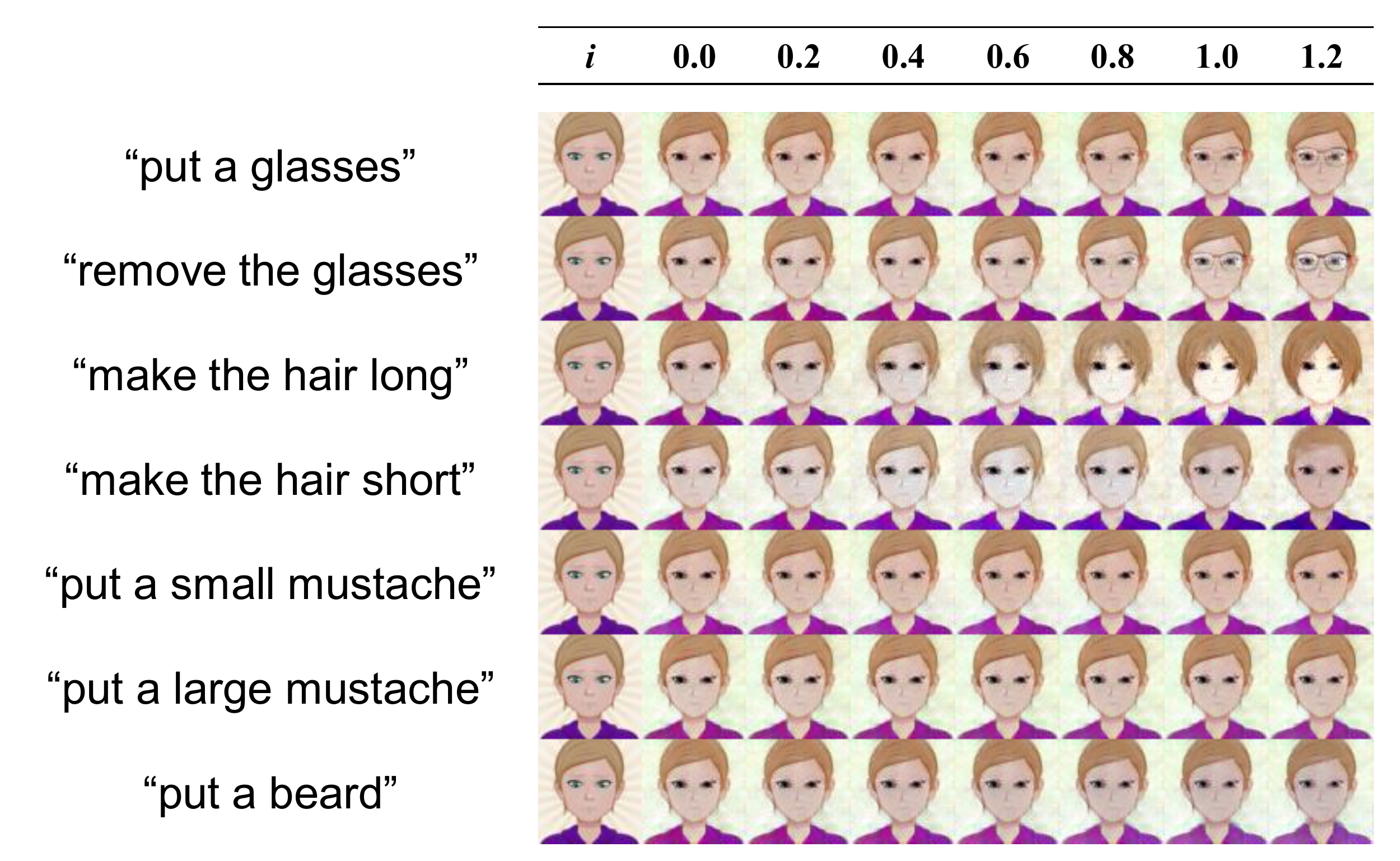}
\vspace{0mm}
\caption{\label{fig:1}Avatar generation example~1.}
\vspace{0mm}
\end{figure}

\begin{figure}[t]
\centering
\vspace{-10mm}
\includegraphics[width=1.0\textwidth]{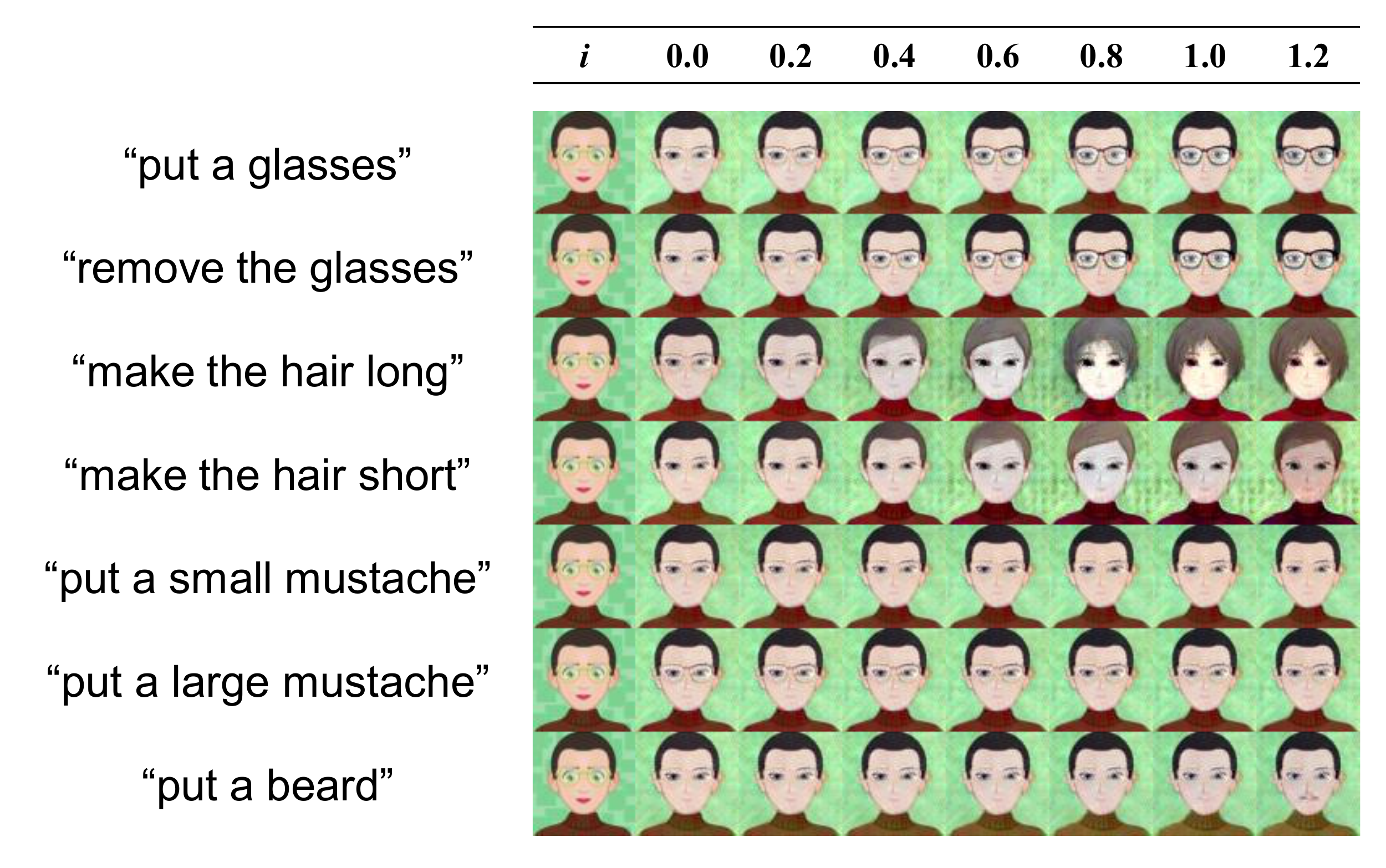}
\vspace{0mm}
\caption{\label{fig:5}Avatar generation example~5.}
\vspace{0mm}
\end{figure}

\begin{figure}[t]
\centering
\vspace{-20mm}
\includegraphics[width=1.0\textwidth]{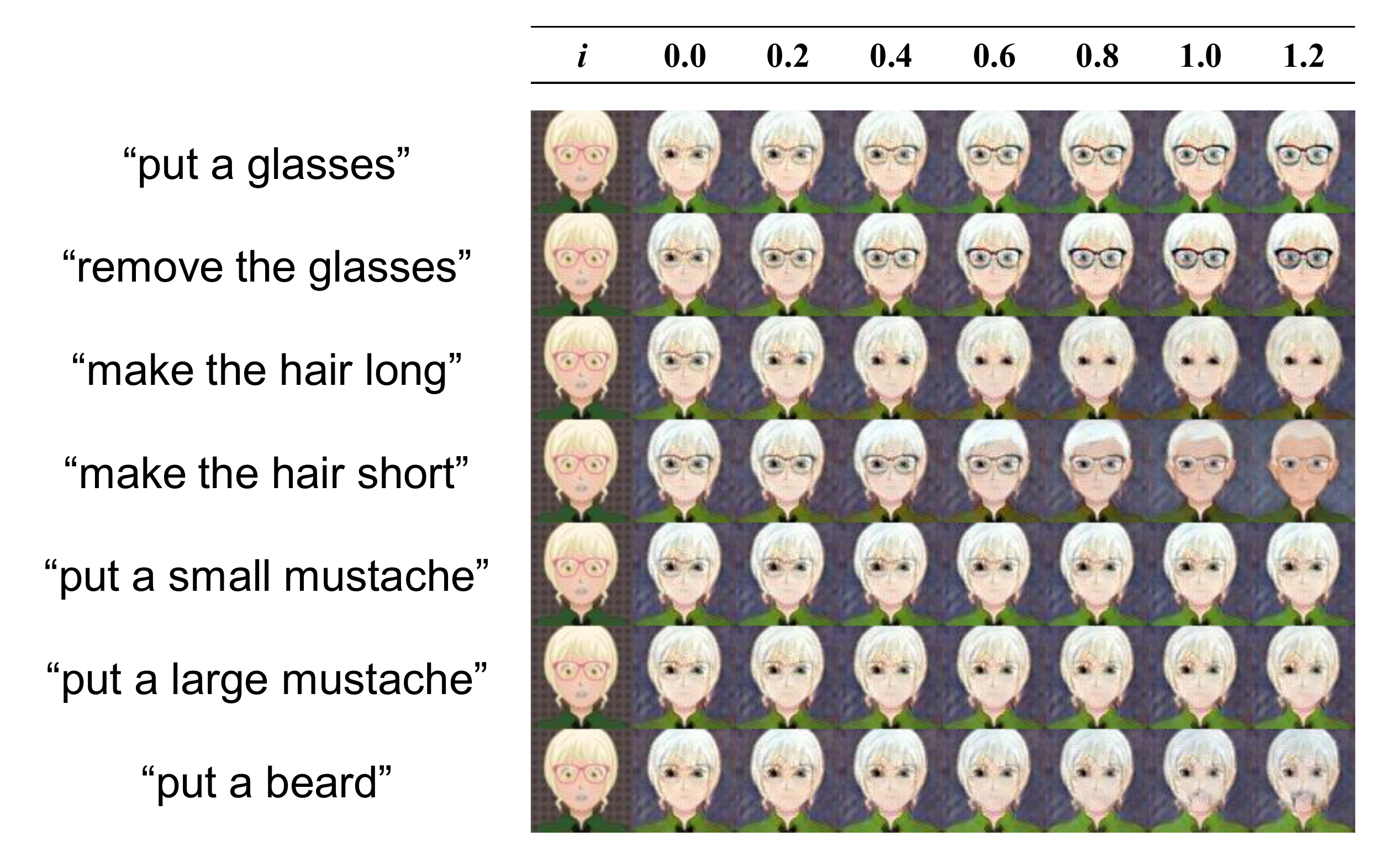}
\vspace{0mm}
\caption{\label{fig:6}Avatar generation example~6.}
\vspace{-20mm}
\end{figure}

Fig~\ref{fig:1} to Fig~\ref{fig:6} show the generation results. First left column indicates the same source image and interpolation generation results are rightward. $i$ means the coefficient of instruction vector, namely we modified (3) to  $\phi_{fc} = FC(\phi_{im},i \cdot \phi_{i}^T)$ in generation. 
 We found that the generation of significant change tends to be successful. For example, natural language instruction captured the concept of putting a glasses, making the hair long or short, putting beard. Notably, the bottom row of Fig~\ref{fig:5} and Fig~\ref{fig:6} shows the beard appeared even though the woman image with the beard does not exist on the whole dataset. However, small changes such as changing mouth, eyes, nose, ears are not well learned in our model. Moreover, the second row, ``remove the glasses" failed. We think it is because the word ``glasses" affected strongly than the verb, ``put" or ``remove".
 
\clearpage

\begin{figure}[t]
\centering
\vspace{-20mm}
\includegraphics[width=1.0\textwidth]{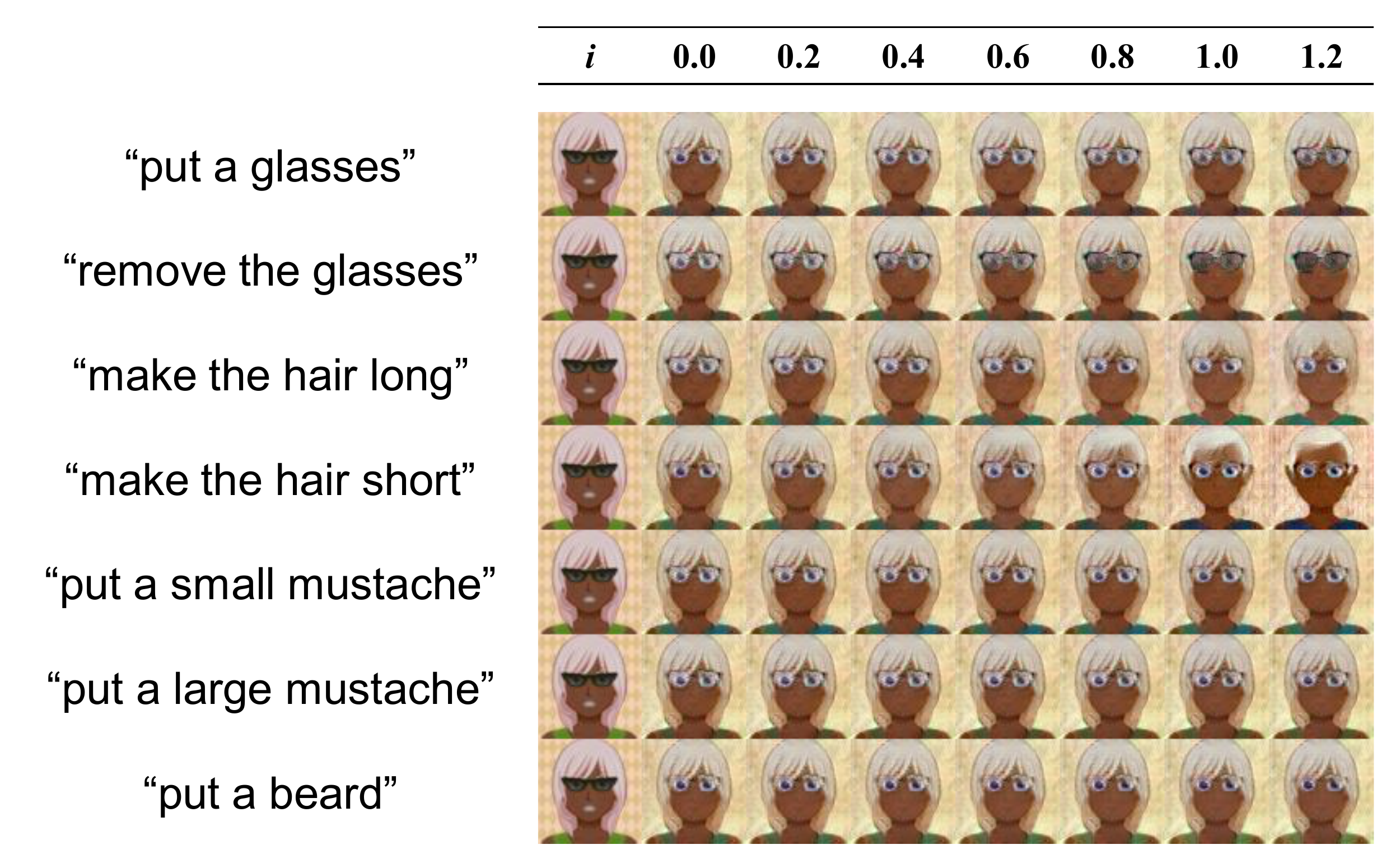}
\vspace{0mm}
\caption{\label{fig:2}Avatar generation example~2.}
\vspace{-20mm}
\end{figure}

\begin{figure}[t]
\centering
\vspace{-10mm}
\includegraphics[width=1.0\textwidth]{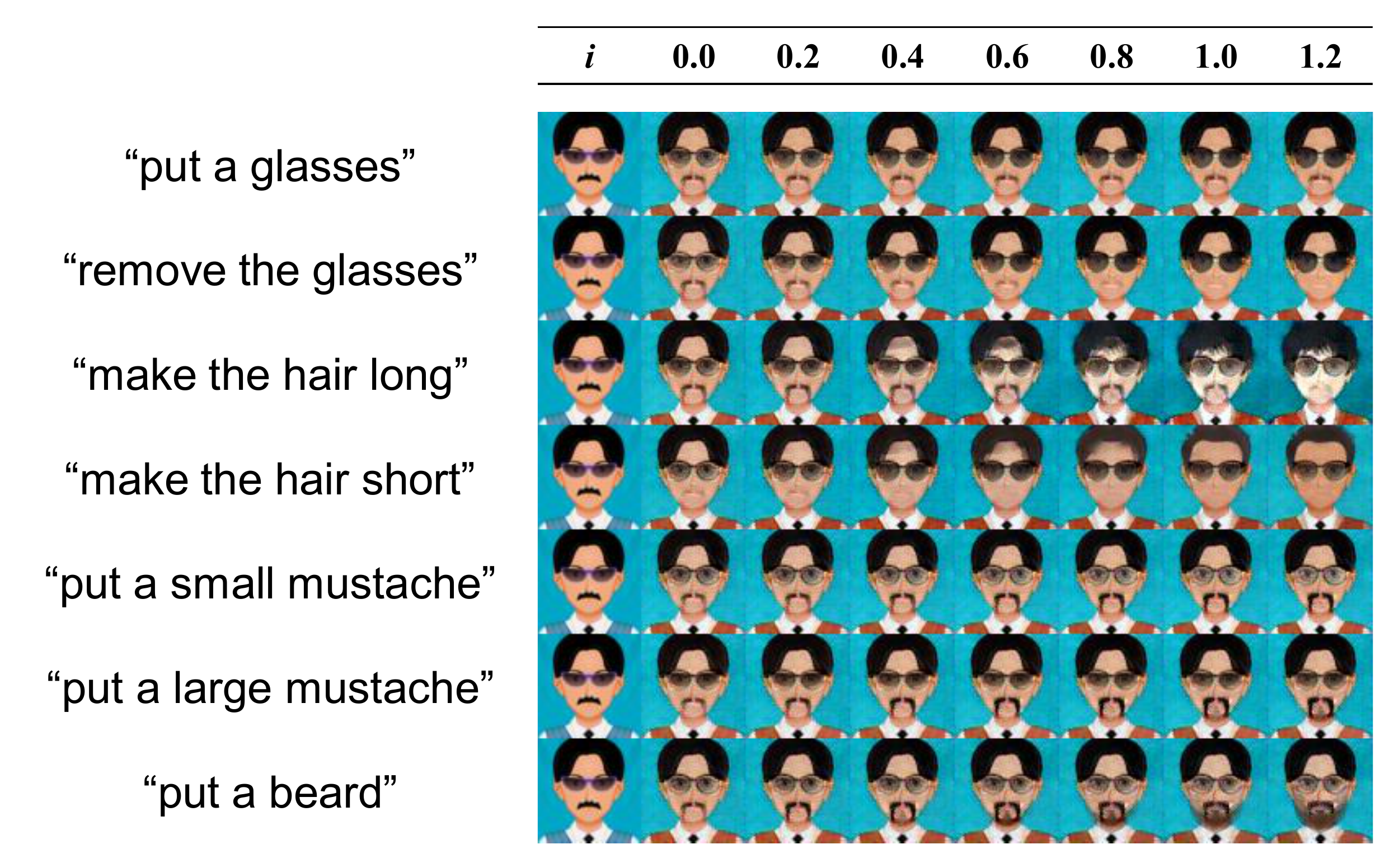}
\vspace{0mm}
\caption{\label{fig:3}Avatar generation example~3.}
\vspace{0mm}
\end{figure}

\begin{figure}[t]
\centering
\vspace{-20mm}
\includegraphics[width=1.0\textwidth]{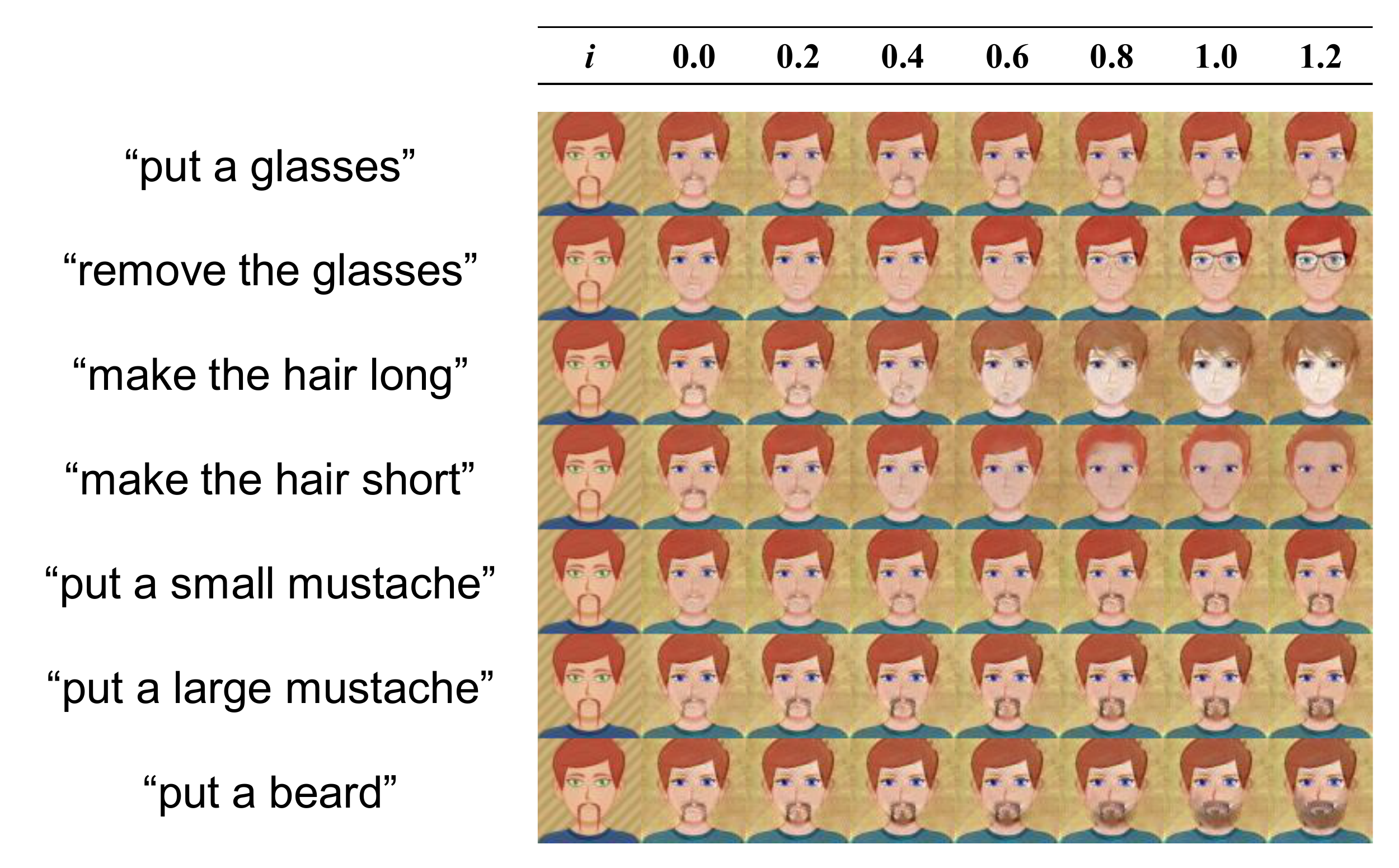}
\vspace{0mm}
\caption{\label{fig:4}Avatar generation example~4.}
\vspace{-20mm}
\end{figure}

\end{document}